\documentclass{article}
\usepackage{amsmath,amssymb,amsfonts}
\usepackage{arxiv}
\usepackage{algorithm}
\usepackage{algorithmic}
\usepackage[utf8]{inputenc} 
\usepackage[T1]{fontenc}    
\usepackage{hyperref}       
\usepackage{url}            
\usepackage{booktabs}       
\usepackage{nicefrac}       
\usepackage{microtype}      
\usepackage{lipsum}		
\usepackage[numbers]{natbib}
\usepackage{doi}
\usepackage{graphicx}   
\usepackage{caption}    
\usepackage{cleveref}   
\usepackage{array}
\usepackage{tabularx}
\usepackage{adjustbox}
\usepackage{enumitem}
\usepackage{comment}

\newcommand{\figref}[1]{\hyperref[#1]{Figure\ref*{#1}}}

\newcommand{\mytabref}[1]{\hyperref[#1]{Table\ref*{#1}}}

\newcommand{\myalgref}[1]{\hyperref[#1]{Algorithm\ref*{#1}}}

\newcommand{\myeqref}[1]{\hyperref[#1]{Equation\ref*{#1}}}

\title{Enhanced Prediction of Ventilator-Associated Pneumonia in Patients with Traumatic Brain Injury Using Advanced Machine Learning Techniques}

\author{ \href{https://orcid.org/0009-0003-8414-2996}{\includegraphics[scale=0.06]{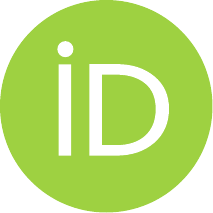}\hspace{1mm}Negin Ashrafi} \\
	Department of Industrial and Systems Engineering\\
	University of Southern California\\
	Los Angeles, CA 90089 \\
	\texttt{ashrafin@usc.edu} \\
	\And
	\href{https://orcid.org/0009-0007-1387-0995}{\includegraphics[scale=0.06]{orcid.pdf}\hspace{1mm}Armin Abdollahi} \\
	Department of Electrical and Computer Engineering\\
	University of Southern California\\
	Los Angeles, CA 90089 \\
	\texttt{arminabd@usc.edu} \\
	\AND
	\href{https://orcid.org/0009-0003-7159-3245}{\includegraphics[scale=0.06]{orcid.pdf}\hspace{1mm}Maryam Pishgar} \\
	Department of Industrial and Systems Engineering\\
	University of Southern California\\
	Los Angeles, CA 90089 \\
	\texttt{pishgar@usc.edu} \\
}

\begin{document}
\maketitle
\begin{abstract}

\noindent\textit{Background:} Ventilator-associated pneumonia (VAP) in traumatic brain injury (TBI) patients poses a significant mortality risk and imposes a considerable financial burden on patients and healthcare systems. Timely detection and prognostication of VAP in TBI patients are crucial to improve patient outcomes and alleviate the strain on healthcare resources.

\noindent\textit{Methods:} We implemented six machine learning models using the MIMIC-III database. Our methodology included preprocessing steps including feature selection with CatBoost and expert opinion, addressing class imbalance with Synthetic Minority Oversampling Technique (SMOTE), and rigorous model tuning through 5-fold cross-validation to optimize hyperparameters. Key models evaluated included SVM, Logistic Regression, Random Forest, XGBoost, ANN, and AdaBoost. Additionally, we conducted SHAP analysis to determine feature importance and performed an ablation study to assess feature impacts on model performance.

\noindent\textit{Results:} XGBoost outperformed the baseline models and the best existing literature. We used metrics, including AUC, Accuracy, Specificity, Sensitivity, F1 Score, PPV, and NPV. XGBoost demonstrated the highest performance with an AUC of 0.940 and an Accuracy of 0.875, which are 23.4\% and 23.5\% higher than the best results in the existing literature with an AUC of 0.706 and an Accuracy of 0.640, respectively. This enhanced performance underscores the models' effectiveness in clinical settings.

\noindent\textit{Conclusions:} This study enhances the predictive modeling of VAP in TBI patients, improving early detection and intervention potential. Refined feature selection and advanced ensemble techniques significantly boosted model accuracy and reliability, offering promising directions for future clinical applications and medical diagnostics research.

\end{abstract}



\maketitle

\section{Background}
Traumatic brain injury (TBI) affects approximately 250 per 100,000 individuals globally, contributing to 30–50\% of trauma-related mortalities, with adolescents, young adults, and older adults being the most affected groups \cite{ref1,ref2}. TBI occurs from forceful impacts to the head or body, resulting in varying degrees of cognitive and physical impairment \cite{ref3}. Complicating the management of TBI is the frequent occurrence of ventilator-associated pneumonia (VAP), a lung infection that develops in patients requiring mechanical ventilation. VAP not only exacerbates the morbidity and mortality associated with TBI but also prolongs hospital stays and increases healthcare costs \cite{ref4}. Therefore, early identification and prediction of VAP in TBI patients are crucial for improving patient outcomes and reducing the burden on healthcare systems.

Several studies have investigated the prediction and management of VAP in TBI patients, leveraging diverse methodologies and datasets to enhance our understanding of this complex clinical scenario. Wang et al. utilized machine learning techniques to predict VAP in patients with TBI, leveraging data from the Medical Information Mart for Intensive Care III (MIMIC-III) database. Their study explored various models, including XGBoost, SVM, Logistic Regression, and Random forest, emphasizing the importance of identifying pertinent features for accurate prediction \cite{ref5}. Robba et al., on the other hand, conducted a large-scale prospective observational study to analyze the incidence, risk factors, and outcomes of VAP in TBI patients. Their findings underscored the multifactorial nature of VAP development, highlighting age and smoking as potential risk factors influencing patient susceptibility \cite{ref6}.

Luo et al. investigated the impact of VAP on the prognosis of ICU patients within 90 and 180 days, identifying critical factors such as diabetes, length of ICU stay, and COPD that influence patient outcomes. Their study also proposed methodologies for enhancing diagnostic approaches in identifying VAP patients \cite{ref7}. Recent studies have shown significant improvements in the power of predicting mortality in mechanically ventilated ICU patients using advanced techniques and enhanced feature selection, demonstrating the effectiveness of deep learning and machine learning models in clinical predictions \cite{ref23}.

Collectively, these studies underscore the importance of predictive modeling, risk factor analysis, and feature selection in improving the identification and management of VAP in TBI patients. Machine learning algorithms offer promising avenues for predicting VAP development while understanding risk factors and prognosis aids in personalized patient care. Further research integrating diverse methodologies and datasets is warranted to enhance the accuracy and generalizability of predictive models in this critical domain.

\section{Methodologies}
\subsection{Data Source and Inclusion Criteria}
In this study, we employed the MIMIC-III database, a publicly available clinical dataset that includes detailed medical information from over 40,000 patients \cite{ref9}. These patients were treated at two prominent hospitals in the United States between 2001 and 2012. The dataset encompasses over 60,000 hospital admissions and more than 20,000 intensive care unit (ICU) stays, amounting to more than 40 million individual clinical record entries. These entries cover a wide range of data points, including physiological signals, medication administrations, laboratory test results, and clinical notes. The extensive scope and detailed nature of MIMIC-III make it an invaluable resource for conducting rigorous biomedical research, enabling us to thoroughly analyze and validate our methodologies within a clinical setting.

\subsection{Patient Extraction}

\begin{figure}[!htb]
    \centering
    \includegraphics[width=0.35\linewidth]{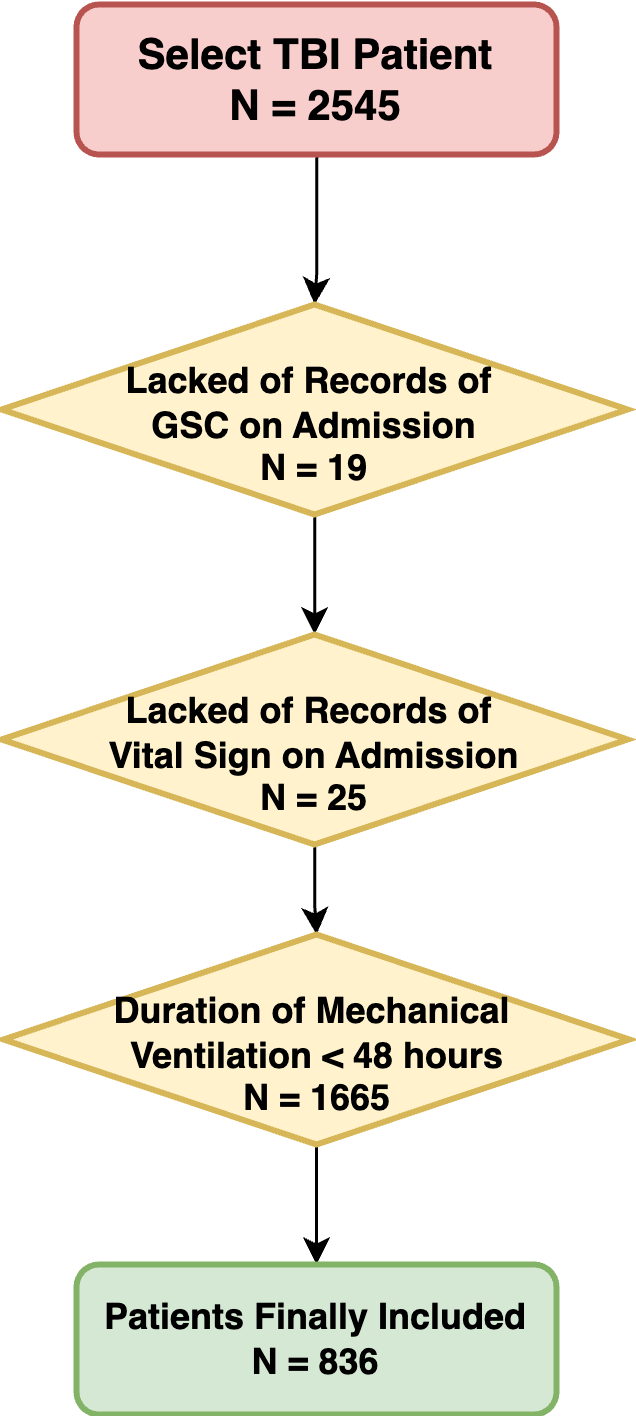}
    \captionsetup{justification=centering}
    \caption{Flow diagram of the patient selection process}
    \label{fig:1}
\end{figure}

To identify Traumatic Brain Injury (TBI) patients for our study, specific International Classification of Diseases and Ninth Revision (ICD-9) codes were utilized. Patients were initially selected based on the following codes: 80,000–80,199; 80,300–80,499; 8500–85419, which identified an initial cohort of 2,545 patients \cite{ref10}. Exclusions were applied for patients lacking Glasgow Coma Scale (GCS) records at admission (19 patients) or those without recorded vital signs at admission (25 patients). Further exclusions were made for patients who underwent mechanical ventilation for less than 48 hours, impacting 1,665 patients. After all exclusions, the final cohort consisted of 836 patients. Within this group, 328 were identified as positive for Ventilator-Associated Pneumonia (VAP), while 508 were negative. This methodology ensures a focused examination of the impact of TBI on the risk of VAP, guided by our predefined clinical criteria. \figref{fig:1} shows the patient extraction process

\subsection{\textit{Feature Selection}}
In our study, the target variable, Ventilator-Associated Pneumonia (VAP), is identified through a structured diagnostic approach that encompasses three key criteria: radiologic, systemic, and pulmonary signs. (1) For radiologic confirmation, a patient must show at least one of the following: new or progressive and persistent infiltrate, consolidation, or cavitation on lung imaging, indicative of the physical manifestations of pneumonia. (2) Systemically, the criteria require either a fever exceeding 38°C or an abnormal white blood cell count, with thresholds set below 4,000/mL or above 12,000/mL, signaling an immune response to infection. (3) On the pulmonary front, a diagnosis is supported by the presence of at least two symptoms: purulent sputum, deteriorating gas exchange, and worsening of respiratory symptoms such as cough, dyspnea, tachypnea, or new breath sounds \cite{ref11}. This multidimensional diagnostic framework ensures a thorough and precise identification of VAP in critically ill patients, capturing the complex clinical profile necessary for accurate diagnosis and subsequent treatment planning. This is summarized in \myalgref{alg:VAP_Diagnosis}.

\begin{algorithm}
\caption{Diagnostic Algorithm for Identifying VAP}
\label{alg:VAP_Diagnosis}
\begin{algorithmic}
\vspace{-1pt}
\STATE \textbf{Input:} Patient's clinical data
\vspace{-1pt}
\STATE \textbf{Output:} VAP diagnosis
\vspace{-1pt}

\STATE \textbf{Step 1: Radiologic Confirmation (RC)}
\IF{Patient symptoms \(\in\) (infiltrate $\lor$ consolidation $\lor$ cavitation)}
    \STATE RC = True
\ELSE
    \STATE RC = False
\ENDIF
\STATE \textbf{Step 2: Systemic Confirmation (SC)}
\IF{((fever $>$ 38°C) $\lor$ (white blood cell count $<$ 4000/mL) $\lor$ (white blood cell count $>$ 12000/mL))}
\vspace{-1pt}
    \STATE SC = True
\ELSE
    \STATE SC = False
\ENDIF
\STATE \textbf{Step 3: Pulmonary Confirmation (PC)}
\vspace{-1pt}
\STATE Initialize symptom\_count = 0
\FOR{each symptom in \{purulent sputum, deteriorating gas exchange, excessive cough, excessive dyspnea, excessive tachypnea, new breath sounds\}}
    \IF{Patient has symptom}
        \STATE symptom\_count = symptom\_count + 1
    \ENDIF
\ENDFOR
\IF{symptom\_count $\geq$ 2}
    \STATE PC = True
\ELSE
\vspace{-1pt}
    \STATE PC = False
\ENDIF
\IF{RC $\land$ SC $\land$ PC}
\vspace{-1pt}
    \STATE VAP = True
\ELSE
    \STATE VAP = False
    \vspace{-1pt}
\ENDIF

\end{algorithmic}
\end{algorithm}

37 candidate features were meticulously chosen to capture a broad spectrum of clinical information crucial for assessing the patients' health outcomes, which is shown in \mytabref{tab:1}. These features are categorized into demographic details, disease-related attributes, vital signs at admission, intracranial injury classifications, laboratory tests, medical interventions, and other relevant clinical data. The demographic features include age, gender, and ethnicity, which are foundational for adjusting clinical analyses. Disease-related features such as smoking history, diabetes, hypertension, and other significant conditions are included due to their potential impact on patient prognosis. Vital signs such as heart rate and blood pressure, recorded at admission, provide immediate clinical context. Intracranial injury features, including conditions like subarachnoid hemorrhage and subdural hematoma, are particularly critical for patients with traumatic brain injuries. Laboratory tests, including measurements of platelet count, hemoglobin levels, and blood urea nitrogen, among others, offer detailed biochemical insights that are indispensable for monitoring patient health and response to treatment. Medical interventions like tracheostomy and neurosurgery, recorded to capture the intensity and nature of the medical care provided, also contribute to understanding patient outcomes. The selection process was meticulously guided by consultations with a clinical expert and a comprehensive review of the existing literature. This rigorous approach ensured that each feature included in the study was highly relevant to the clinical questions posed, as a result significantly enhancing the robustness and practical applicability of our predictive models. This thoughtful compilation of features enables a holistic view of patient health, supporting the identification of patterns and predictors of recovery in a clinically diverse population.

\begin{table*}[t]
\centering
\renewcommand{\arraystretch}{1.6}
\caption{List of candidate features categorized by type, including vital signs, laboratory tests, demographics, Intracranial injury, disease-related features, and medical interventions.}
\label{tab:1}
\begin{tabularx}{\textwidth}{|l|X|l|X|}
\hline
\textbf{Feature Type} & \textbf{Feature Name} & \textbf{Feature Type} & \textbf{Feature Name} \\
\hline
\textbf{Vital signs} & Diastolic blood pressure & \textbf{Laboratory tests} & Blood urea nitrogen \\
 &  Respiratory rate & & Red blood cell \\
 & Systolic blood pressure & & Hemoglobin \\
 & Heart rate & & Glucose \\
\cline{1-2}
\textbf{Demographics} & Age & & Platelet \\
 & Gender & & Serum sodium \\
 & Ethnicity & & INR \\
\cline{1-2}
\textbf{Disease related features} & Chronic renal disease & & Serum chloride \\
 & Chronic liver disease & & Anion gap \\
 & Hypertension & & Serum potassium \\
\cline{3-4}
 & Previous myocardial infarction & \textbf{Medical interventions features} & Anticoagulant within 24h \\
 & Cancer & & Tracheostomy \\
 & Smoking history & & Neurosurgery \\
 & Cerebral vascular disease & &PEG \\
 & Diabetes & & Parenteral nutrition\\
\cline{1-2}
\textbf{Intracranial injury features} & Subarachnoid hemorrhage & & Platelet transfusion within 24h  \\
 & Subdural hematoma & & RBC transfusion within 24h \\
\cline{1-2}
\textbf{Other clinical features} & 30 day mortality & & \\
 & Hospital stay length & & \\
 & ICU stay length & & \\
\hline
\end{tabularx}
\end{table*}

\begin{figure*}[!htb]
    \centering
    \includegraphics[width=0.95\linewidth]{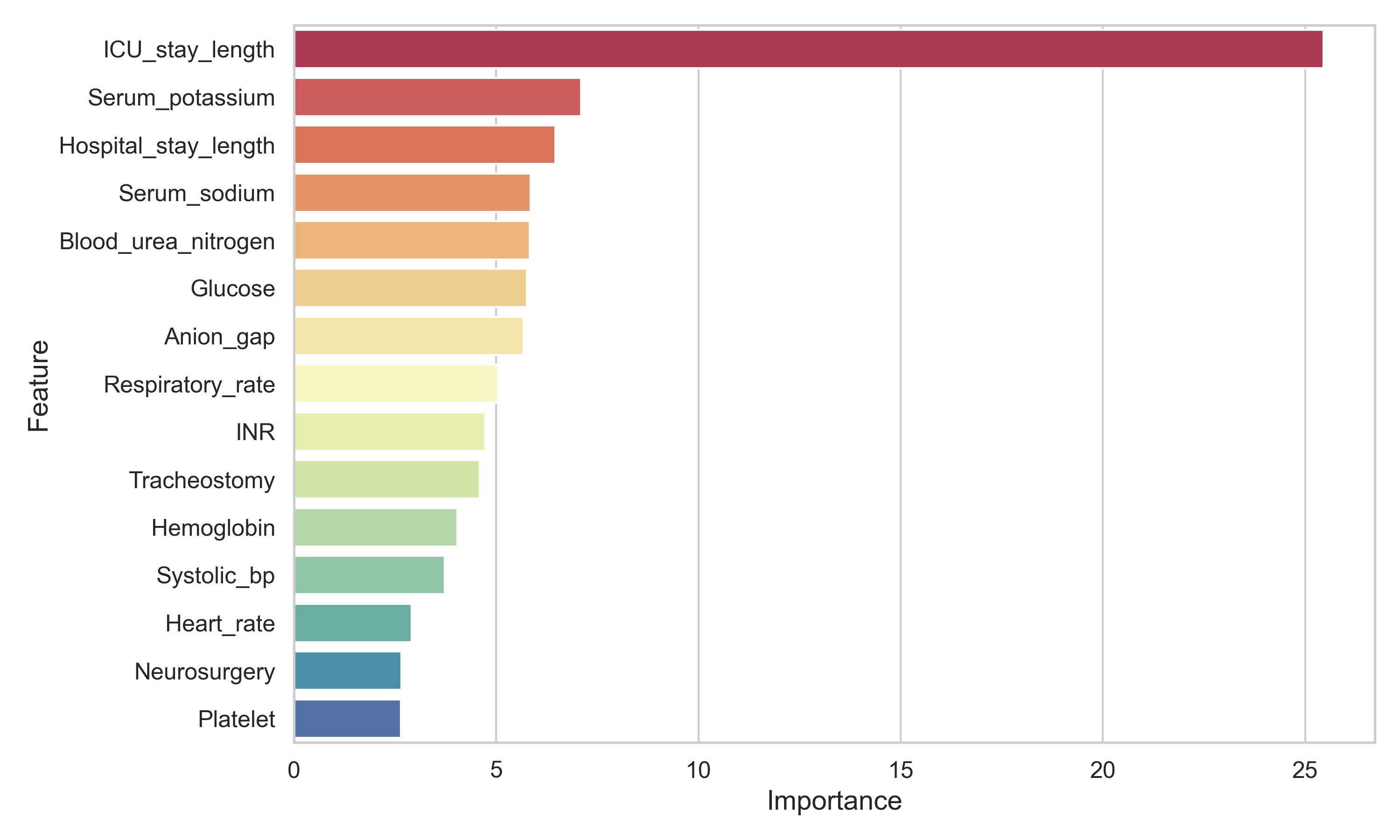}
    \captionsetup{justification=centering}
    \caption{Top 15 features based on CatBoost feature importance scores, highlighting the most impactful ones.}
    \label{fig:2}
\end{figure*}

To refine the feature set, we first applied a correlation matrix to identify and remove highly correlated features, thereby reducing redundancy within our dataset. This process led to the exclusion of features such as PEG (percutaneous endoscopic gastrostomy), serum chloride, and red blood cell count, which were found to provide overlapping information. Subsequently, we employed the CatBoost algorithm to determine the relative importance of the remaining features, guiding us to focus on those most influential for our predictive model. The top 15 features, as identified by CatBoost, include critical indicators such as ICU stay length, serum potassium, and hospital stay length, among others, as shown in the accompanying feature importance chart, which is shown in \figref{fig:2}. These features were selected for their strong predictive power and clinical relevance, based on expert opinion, ensuring that our model is both accurate and interpretable in assessing patient outcomes.

\subsection{\textit{Data Preprocessing }}

In the preprocessing stage of our data analysis, we implemented systematic approaches to manage missing values, encode categorical data, and scale features appropriately. For numerical data, missing values were imputed with the median of each respective feature, given its robustness to outliers, thereby maintaining the integrity of the data distribution. For categorical data, missing entries were filled using the mode, ensuring the most frequent category was used for a consistent and representative fill. Further enhancing our model's ability to process categorical variables, one-hot encoding was applied. 

In terms of feature scaling, different strategies were employed based on the specific requirements of the algorithms used in subsequent analyses. For models sensitive to the scale of input features, a min-max scaler was utilized to normalize the data within a range of 0 to 1 \cite{ref12}. This scaling preserves the relationships among the original data points. For other algorithms, we applied a standard scaler, which standardized features by removing the mean and scaling to unit variance. This approach is particularly beneficial for models that assume data is normally distributed, such as many linear models, and helps in reducing the influence of outliers.

\begin{figure*}[!htb]
    \centering
    \includegraphics[width=0.7\linewidth]{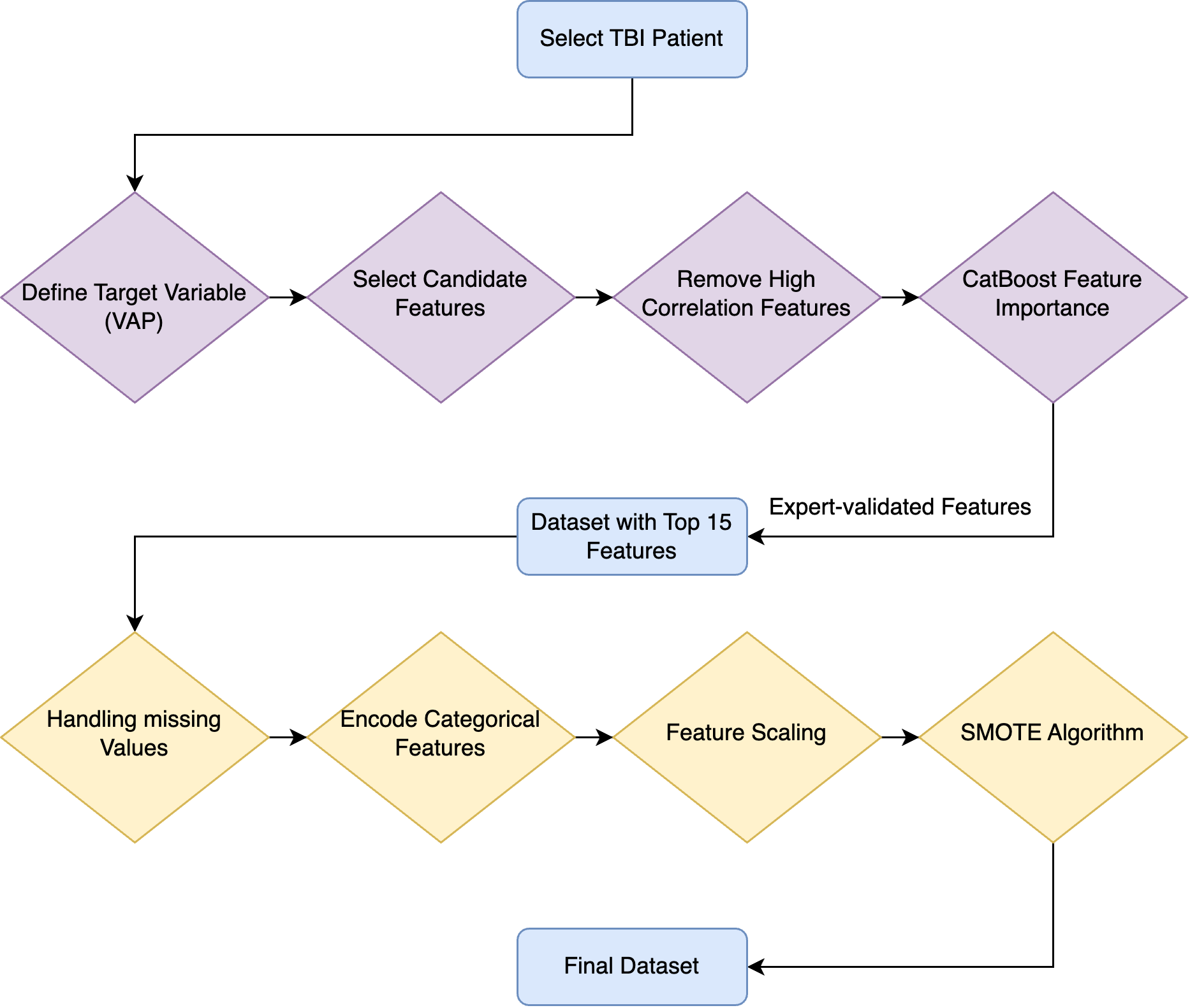}
    \captionsetup{justification=centering}
    \caption{Data preprocessing workflow, illustrating the steps from patient selection to the creation of the final dataset.}
    \label{fig:3}
\end{figure*}

To handle the class imbalance, we integrated the Synthetic Minority Over-sampling Technique (SMOTE) within a 5-fold cross-validation framework, enhancing the representation of the minority class during the training phase \cite{ref13}. SMOTE was applied to the training folds only, synthesizing new samples to augment the minority (positive) class. This ensured that each training set used in the cross-validation was balanced while the original distribution of the classes was preserved in the validation folds to maintain the integrity of the validation process. By doing so, we aimed to improve the model's generalizability and prevent data leakage, allowing us to evaluate the model's performance against unmodified data, thus ensuring more accurate and robust outcomes. \figref{fig:3}  illustrates the workflow for data preprocessing.

\subsection{\textit{Ablation process}}

We implemented a stepwise elimination process to determine if the 15 selected features had an adverse effect on the model's performance. This involved systematically removing variables that had an adverse effect on the model's performance, which we measured by calculating the 95\% confidence interval (CI) of the Area Under the Receiver Operating Characteristic Curve (AUROC). We started by calculating the baseline AUROC using all 15 features.

In each iteration, we temporarily removed one feature from the current set and recalculated the AUROC. If the AUROC for the new set of features showed improvement, we updated the current feature set by permanently excluding the feature and recorded the improved AUROC. This iterative process continued, with the model being re-evaluated after each removal, until no further improvements were observed\cite{ref29}.

Ultimately, all 15 features were found to positively influence the model's performance, leading us to retain the entire set. This process ensured the final model included only features that contributed to its predictive power. The details are provided in the results section.


\subsection{\textit{Modeling}}

After data preprocessing, we used the top 15 important features for the modeling part. Then, we divided the dataset into training and testing sections \cite{ref14}. 70\% of the data was used for model training and 30\% of the data for testing.
Six models have been implemented in this project, including Support Vector Machine (SVM), Logistic Regression (LR), Random Forest (RF), XGBoost, Artificial Neural Network (ANN), and AdaBoost. A 5-fold cross-validation method has been utilized for model tuning to choose the best pair of Hyper-parameters of each model by using the training data, to ensure the performance of the models.

For the SVM, we adjusted the regularization parameter, kernel type, and kernel parameters to manage model complexity and fit \cite{ref15}. In LR, we tuned the regularization strength and L1 and L2 penalty types to balance bias and variance, enhancing generalizability \cite{ref16, ref27}. The RF model was optimized by setting the number of trees, the maximum depth of the trees, and the quality criterion for splits to improve performance across different data subsets \cite{ref17, ref28}. In the XGBoost model, adjustments included the colsample\_bytree, learning rate, max\_depth, min\_child\_weight, n\_estimators, reg\_alpha, reg\_lambda, scale\_pos\_weight, and subsample to increase accuracy and prevent overfitting. The colsample\_bytree parameter specifies the fraction of features to be randomly sampled for each tree, helping to prevent overfitting. The learning rate controls the step size for each boosting iteration, balancing between learning quickly and preventing overfitting. Max\_depth sets the maximum depth of a tree, impacting model complexity and overfitting potential. Min\_child\_weight determines the minimum sum of instance weight (hessian) needed in a child, providing regularization by preventing overly specific trees. N\_estimators is the number of boosting rounds. The reg\_alpha and reg\_lambda are L1 and L2 regularization terms, respectively, which add penalties to the model to prevent overfitting. Scale\_pos\_weight balances the positive and negative weights, useful for handling class imbalance. Subsample is the fraction of samples to be used for fitting the individual base learners, which helps prevent overfitting by introducing randomness. These adjustments optimize model performance and robustness against overfitting \cite{ref25, ref30, ref22}.

The artificial neural network (ANN) consisted of an input layer, one hidden layer, and an output layer, utilizing ReLU (Rectified Linear Unit) activation functions for the input and hidden layers to facilitate non-linear learning and a sigmoid activation function in the output layer for binary classification probabilities \cite{ref19, ref26, ref8}. Lastly, the AdaBoost model was fine-tuned by adjusting the depth of the base estimators, the learning rate, and the number of weak learners to enhance the fit and robustness of training data \cite{ref20}.


From a data perspective, using metrics such as AUC and sensitivity, it is evident that the proposed model not only performs better overall but also exhibits a superior ability to predict the minority class. The model demonstrating the highest performance on the test set was selected as the best. We also assessed our models' performance by calculating accuracy, precision, F1-score, and specificity.  Details will be discussed in the results section.

\section{RESULTS}

\subsection{\textit{Statistical Analysis between Cohorts}}

The dataset under examination divided patients into two cohorts: the training cohort (comprising 585 patients) and the test cohort (comprising 251 patients). These cohorts were selected to validate the accuracy and reliability of clinical predictions or treatment outcomes using statistical models.

For continuous variables, such as ICU stay length, serum sodium, and glucose levels, the differences between cohorts were assessed using two-sided t-tests. This test choice is appropriate under the assumption that the data approximately follow a normal distribution, though it is robust to mild deviations from this assumption. For binary or categorical variables like Tracheostomy and Neurosurgery, a Chi-Square test was employed to determine if the distribution of categories differs significantly between the cohorts \cite{ref21}. This test is chosen assuming that the observed frequencies are large enough to meet the test’s conditions.

The primary aim of this analysis is to evaluate whether there are statistically significant differences between the training and test cohorts across various clinical metrics. By understanding these differences, the study aims to ensure that the training data are representative of the test data, thereby validating the model's applicability to future, unseen patients. \mytabref{tab:cohort_comparison} illustrates a comparison of the feature values between train and testing cohorts by capturing the mean and standard deviation of features in each cohort along with the p-value associated with these two cohorts. the p-value of 0.05 is chosen as the threshold for observing significant differences between the cohorts.

\begin{table}[ht]
\caption{Cohort Comparison of Training and Test Sets for all 15 features with their associated p-value.}
\centering
\renewcommand{\arraystretch}{1.2}
\begin{tabular}{|>{\centering\arraybackslash}p{3cm}|>{\centering\arraybackslash}p{4cm}|>{\centering\arraybackslash}p{4cm}|>{\centering\arraybackslash}p{2cm}|}
\hline
\textbf{Feature} & \textbf{Train Mean (Std)} & \textbf{Test Mean (Std)} & \textbf{P-Value} \\
\hline
ICU Stay Length & 6.736 (7.152) & 6.175 (6.487) & 0.286 \\
\hline
Serum Potassium & 3.989 (0.684) & 4.008 (0.717) & 0.723 \\
\hline
Hospital Stay Length & 12.434 (14.028) & 11.182 (10.698) & 0.206 \\
\hline
Serum Sodium & 139.767 (4.119) & 139.234 (4.381) & 0.093 \\
\hline
Blood Urea Nitrogen & 17.800 (10.823) & 17.390 (8.753) & 0.596 \\
\hline
Glucose & 156.784 (50.918) & 152.494 (47.112) & 0.254 \\
\hline
Anion Gap & 15.266 (3.589) & 15.013 (3.423) & 0.344 \\
\hline
Respiratory Rate & 17.629 (5.075) & 17.806 (3.485) & 0.615 \\
\hline
INR & 1.308 (0.466) & 1.402 (1.320) & 0.128 \\
\hline
Tracheostomy & 0.214 (0.410) & 0.199 (0.400) & 0.638 \\
\hline
Hemoglobin & 12.592 (2.169) & 12.501 (2.176) & 0.577 \\
\hline
Systolic BP & 132.213 (14.891) & 131.891 (16.211) & 0.780 \\
\hline
Heart Rate & 87.283 (15.525) & 86.852 (15.534) & 0.713 \\
\hline
Neurosurgery & 0.318 (0.466) & 0.283 (0.451) & 0.314 \\
\hline
Platelet & 232.297 (91.167) & 238.709 (89.538) & 0.349 \\
\hline
\end{tabular}

\label{tab:cohort_comparison}
\end{table}

The training and test cohorts consist of patients characterized by several clinical metrics. Notable measurements include the length of stay in the ICU and hospital, various blood chemistry markers like serum sodium, potassium, and urea nitrogen, as well as vital signs such as heart rate, and systolic blood pressure. Additional specific interventions or conditions, such as tracheostomies and neurosurgery, are also tracked.

The training cohort, larger in number, likely provides a base for developing and training predictive models, with values such as an average ICU stay of approximately 6.74 days and a mean serum sodium level of about 139.77 mmol/L. On the other hand, the smaller test cohort is utilized to evaluate the performance and generalizability of the models developed from the training cohort data. It shows slightly different averages, such as a shorter ICU stay at around 6.17 days and a serum sodium level averaging 139.24 mmol/L.

In both cohorts, metrics such as glucose level and heart rate are monitored to gauge the patient's overall health and immediate medical needs. These measurements, along with others, contribute to a comprehensive profile of each patient, which is crucial for accurate modeling and subsequent decision-making in a clinical setting.

\subsection{\textit{Ablation Study}}
\begin{figure*}[!htb]
    \centering
    \includegraphics[width=0.9\linewidth]{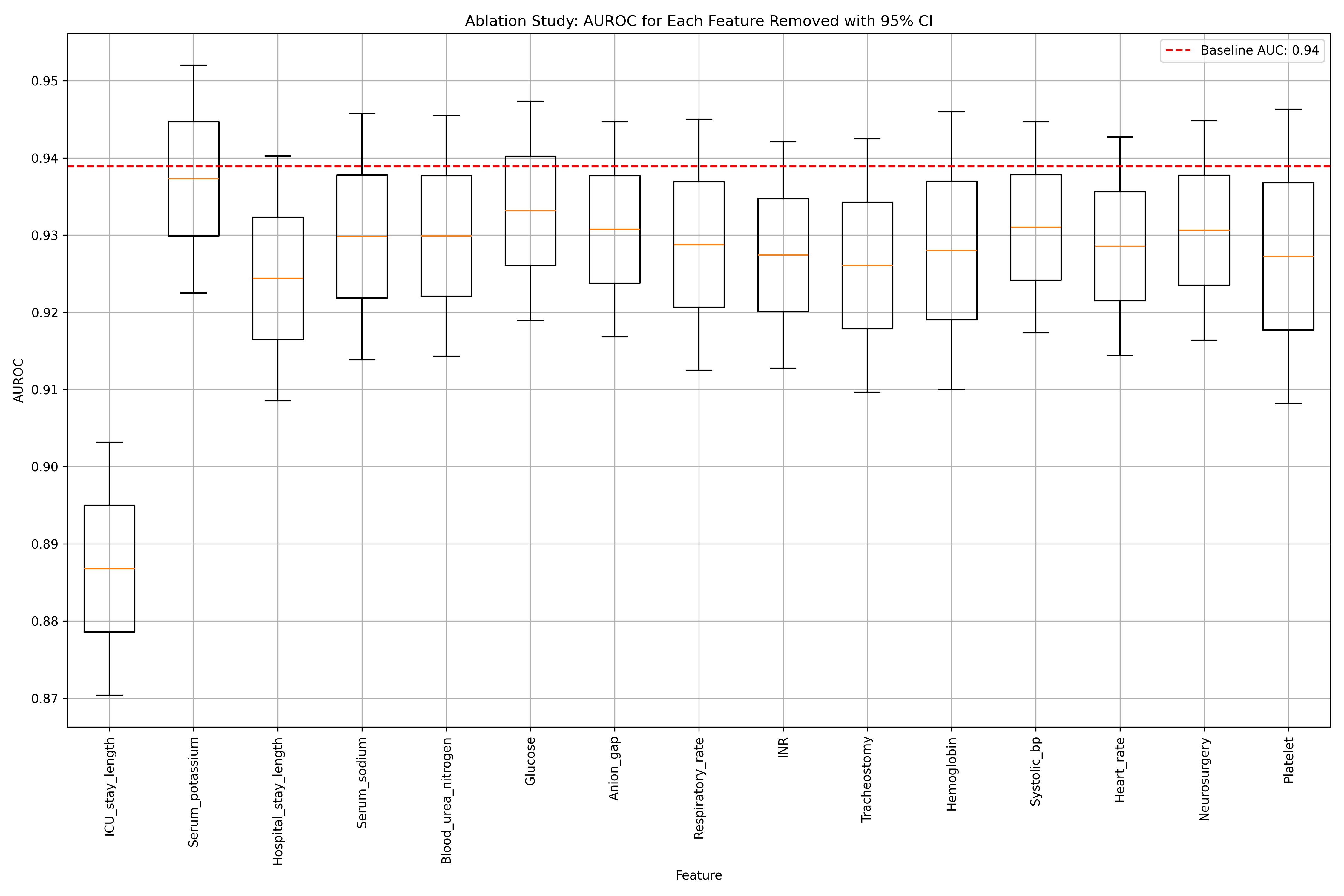}
    \caption{Ablation study for proposed XGBoost model}
    \captionsetup{justification=centering}
    \label{fig:4}
\end{figure*}

The results of our ablation study, comprehensively summarized in \figref{fig:4}, provide a clear and detailed analysis of the model's performance with varying feature sets. This figure suggests that the inclusion of all features is crucial, as any attempt to remove even a single feature results in a noticeable drop in the AUROC. Specifically, the figure demonstrates that the removal of any feature does not lead to an increase in AUROC; rather, it leads to a reduction, indicating that each feature contributes uniquely to the model's predictive accuracy. The baseline model, which incorporates all 15 features, achieves a commendable AUROC of 0.94. The study further reveals that no feature removal leads to an AUROC value higher than this benchmark. This finding underscores the integral role of the current feature set in maintaining the model's optimal performance, thereby negating the necessity for any further deletions or adjustments of features. The robustness and reliability of the baseline model are thus validated, highlighting the synergistic effect of the included features in enhancing the model's predictive capability.

\subsection{\textit{Evaluation results}}

To optimize the performance of our XGBoost model, a grid search was employed to identify the best hyperparameters. The grid search process involved systematically exploring a range of values for each parameter to find the combination that yield the highest model performance on train. \mytabref{tab:grid_search_results} presents the best parameters identified through this process.

\begin{table}[h!]
\caption{Parameters used in XGBoost Algorithm.}
\renewcommand{\arraystretch}{1.6} 
    \centering
    \begin{tabular}{ll}
        \hline
        \textbf{Parameter} & \textbf{Best Value} \\
        \hline
        colsample\_bytree & 0.7 \\
        learning\_rate & 0.01 \\
        max\_depth & 5 \\
        min\_child\_weight & 5 \\
        n\_estimators & 300 \\
        reg\_alpha & 0.1 \\
        reg\_lambda & 2 \\
        scale\_pos\_weight & 2 \\
        subsample & 0.7 \\
        \hline
    \end{tabular}

    \label{tab:grid_search_results}
\end{table}

The results of our study emphasize the predictive performance of various machine learning techniques for VAP in patients with TBI, focusing primarily on the test set. The evaluation metrics are Area Under the Curve (AUC), Accuracy, F1 Score, Specificity, Sensitivity, Positive Predictive Value (PPV), and Negative Predictive Value (NPV). The XGBoost algorithm emerged as the most effective model.

\begin{figure*}[!htb]
    \centering
    \includegraphics[width=0.75\linewidth]{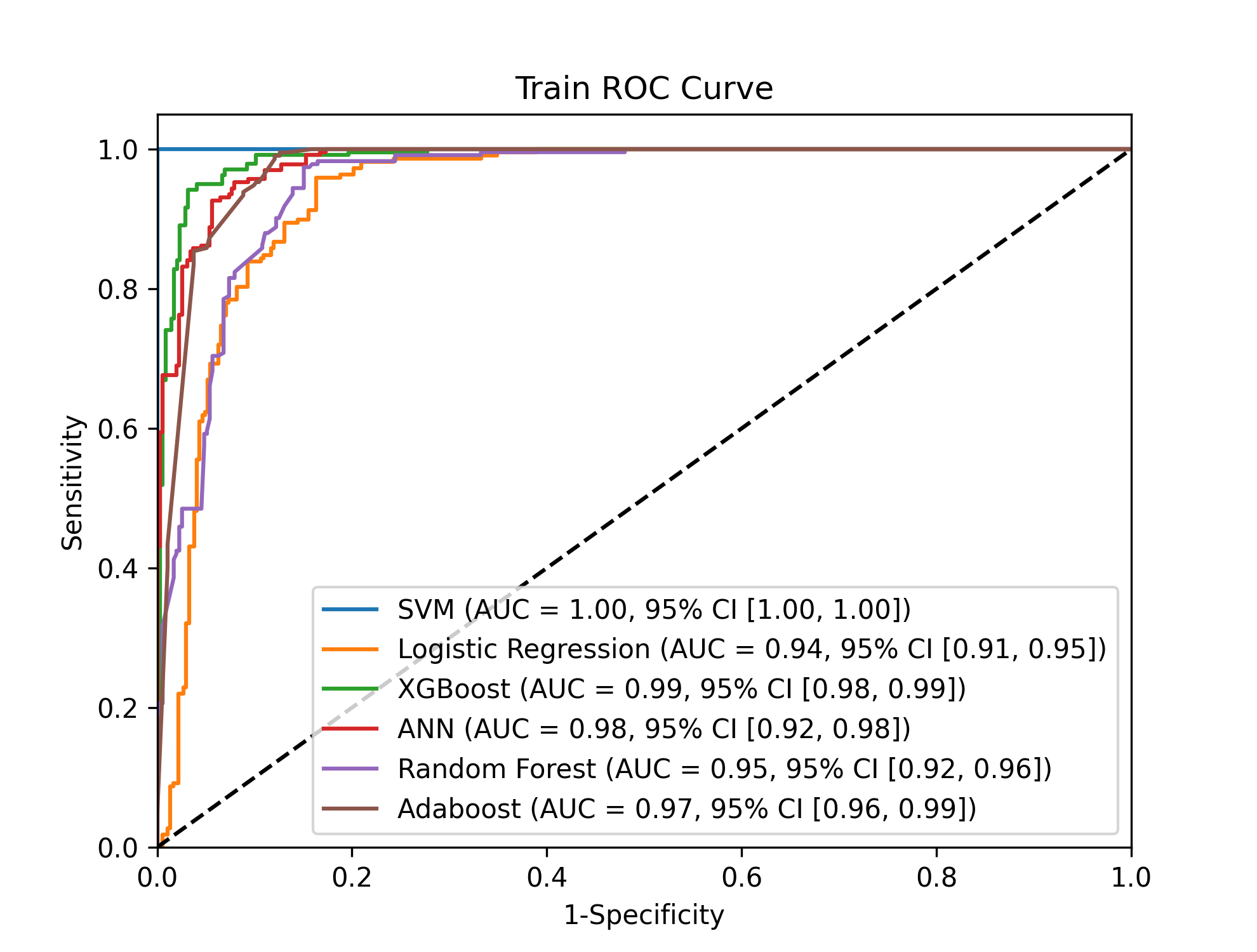}
    \caption{ROC curves of the eight models for the training set. ROC curves of the six models for the test set. SVM, Logistic Regression, XGB, RF, ANN, AdaBoost.
}
    \label{fig:5}
\end{figure*}

As shown in \mytabref{tab:3}, the performance of the models on the training set was initially assessed, with SVM achieving the highest AUC of 1.00. This was further illustrated by the Receiver Operating Characteristic (ROC) curves in \figref{fig:5}, which highlight SVM's exceptional performance on the training data, followed closely by XGBoost with an AUC of 0.99.

\begin{table*}[t]
\centering
\renewcommand{\arraystretch}{1.3}
\caption{Predictive performance of machine learning models for VAP among TBI patients in the training set.}
\label{tab:3}
\scriptsize
\begin{tabularx}{\textwidth}{|>{\centering\arraybackslash}X|>{\centering\arraybackslash}X|>{\centering\arraybackslash}X|>{\centering\arraybackslash}X|>{\centering\arraybackslash}X|>{\centering\arraybackslash}X|>{\centering\arraybackslash}X|>{\centering\arraybackslash}X|}
\hline
\textbf{Models} & \textbf{AUC (95\%CI)} & \textbf{Accuracy (95\%CI)} & \textbf{F1 Score (95\%CI)} & \textbf{Sensitivity (95\%CI)} & \textbf{Specificity (95\%CI)} & \textbf{PPV (95\%CI)} & \textbf{NPV (95\%CI)} \\
\hline
SVM & 0.999 & 0.997 & 0.997 & 0.998 & 0.996 & 0.995 & 0.999 \\
    & (0.998--1.000) & (0.995--1.000) & (0.993--1.000) & (0.996--1.000) & (0.993--1.000) & (0.989--1.000) & (0.997--1.000) \\
\hline
LR & 0.936 & 0.872 & 0.840 & 0.855 & 0.891 & 0.829 & 0.902 \\
    & (0.914--0.949) & (0.851--0.894) & (0.810--0.869) & (0.825--0.884) & (0.867--0.915) & (0.795--0.864) & (0.888--0.915) \\
\hline
XGBoost & 0.988 & 0.915 & 0.904 & 0.984 & 0.866 & 0.834 & 0.989 \\
    & (0.984--0.992) & (0.896--0.935) & (0.879--0.929) & (0.973--0.996) & (0.835--0.896) & (0.796--0.872) & (0.981--0.997) \\
\hline
ANN & 0.980 & 0.875 & 0.829 & 0.811 & 0.911 & 0.866 & 0.891 \\
    & (0.921--0.976) & (0.816--0.935) & (0.735--0.923) & (0.658--0.963) & (0.892--0.929) & (0.834--0.897) & (0.808--0.974) \\
\hline
RF & 0.946 & 0.870 & 0.843 & 0.895 & 0.848 & 0.802 & 0.936 \\
    & (0.920--0.963) & (0.837--0.904) & (0.799--0.886) & (0.809--0.981) & (0.772--0.924) & (0.728--0.876) & (0.887--0.985) \\
\hline
AdaBoost & 0.974 & 0.918 & 0.905 & 0.967 & 0.881 & 0.845 & 0.977 \\
    & (0.963--0.991) & (0.887--0.950) & (0.871--0.939) & (0.934--1.000) & (0.820--0.942) & (0.778--0.912) & (0.955--1.000) \\
\hline
\end{tabularx}
\end{table*}

\begin{figure*}[!htb]
    \centering
    \includegraphics[width=0.75\linewidth]{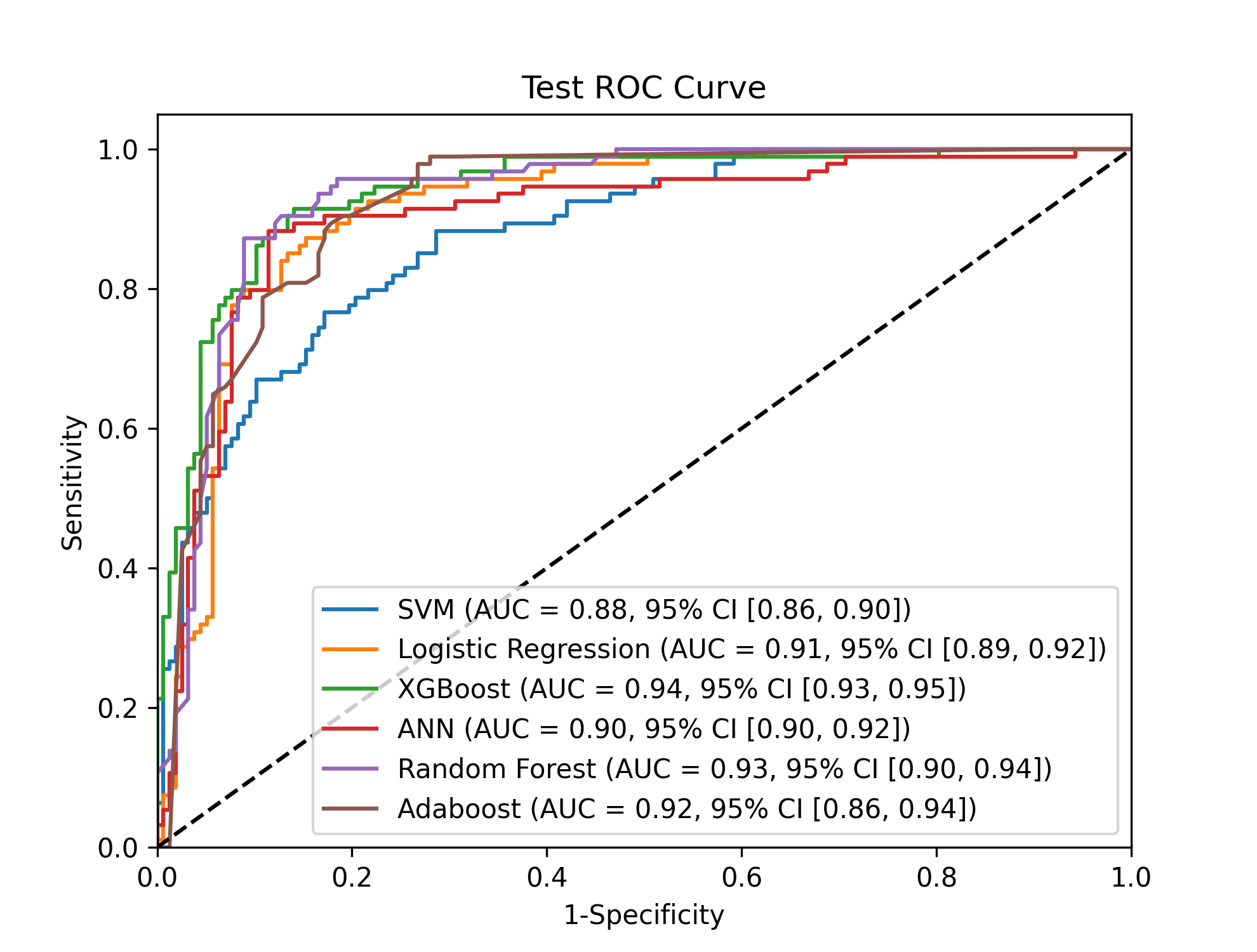}
    \caption{ROC curves of the six models for the test set. ROC curves of the eight models for the test set. SVM, Logistic Regression, XGB, RF, ANN, AdaBoost.
}
    \label{fig:6}
\end{figure*}

However, the true test of a model's generalizability is its performance on unseen data. As shown in \mytabref{tab:4}, XGBoost achieved the highest AUC of 0.940 with a 95\% Confidence Interval (CI) [0.935–0.954] on the test set, demonstrating superior overall performance compared to other models. This was further validated by the ROC curves depicted in \figref{fig:6}, illustrating XGBoost's robust discriminative ability. Additionally, XGBoost maintained a high accuracy of 0.875, sensitivity of 0.896, and specificity of 0.857, indicating its balanced performance across different evaluation metrics.

\begin{table*}[t]
\centering
\renewcommand{\arraystretch}{1.3}
\caption{Predictive performance of machine learning models for VAP among TBI patients in the test set.}
\label{tab:4}
\scriptsize
\begin{tabularx}{\textwidth}{|>{\centering\arraybackslash}X|>{\centering\arraybackslash}X|>{\centering\arraybackslash}X|>{\centering\arraybackslash}X|>{\centering\arraybackslash}X|>{\centering\arraybackslash}X|>{\centering\arraybackslash}X|>{\centering\arraybackslash}X|}
\hline
\textbf{Models} & \textbf{AUC (95\%CI)} & \textbf{Accuracy (95\%CI)} & \textbf{F1 Score (95\%CI)} & \textbf{Sensitivity (95\%CI)} & \textbf{Specificity (95\%CI)} & \textbf{PPV (95\%CI)} & \textbf{NPV (95\%CI)} \\
\hline
SVM & 0.882 & 0.807 & 0.737 & 0.723 & 0.866 & 0.762 & 0.838 \\
    & (0.862--0.903) & (0.773--0.841) & (0.691--0.783) & (0.654--0.793) & (0.822--0.911) & (0.701--0.824) & (0.804--0.873) \\
\hline
LR & 0.914 & 0.847 & 0.784 & 0.745 & 0.908 & 0.828 & 0.856 \\
    & (0.893--0.924) & (0.833--0.860) & (0.762--0.806) & (0.713--0.777) & (0.904--0.911) & (0.818--0.838) & (0.840--0.872) \\
\hline
XGBoost & 0.940 & 0.875 & 0.842 & 0.896 & 0.857 & 0.792 & 0.933 \\
    & (0.935--0.954) & (0.861--0.888) & (0.826--0.858) & (0.877--0.915) & (0.828--0.885) & (0.761--0.823) & (0.922--0.944) \\
\hline
ANN & 0.896 & 0.845 & 0.794 & 0.793 & 0.876 & 0.793 & 0.876 \\
    & (0.895--0.921) & (0.837--0.853) & (0.781--0.806) & (0.777--0.809) & (0.866--0.885) & (0.778--0.808) & (0.867--0.885) \\
\hline
RF & 0.925 & 0.845 & 0.806 & 0.837 & 0.833 & 0.761 & 0.903 \\
    & (0.896--0.935) & (0.798--0.892) & (0.751--0.862) & (0.748--0.926) & (0.755--0.911) & (0.680--0.843) & (0.856--0.949) \\
\hline
AdaBoost & 0.918 & 0.857 & 0.816 & 0.835 & 0.865 & 0.793 & 0.901 \\
    & (0.861--0.936) & (0.833--0.880) & (0.784--0.848) & (0.755--0.915) & (0.822--0.908) & (0.748--0.838) & (0.861--0.942) \\
\hline
\end{tabularx}
\end{table*}

Other model evaluations include Logistic Regression, Artificial Neural Networks (ANN), Random Forest (RF), and AdaBoost. The SVM, despite having the highest training set AUC, showed a lower AUC value of 0.882 in the test set, highlighting the importance of evaluating models on unseen data to ensure their generalizability. Random Forest, Logistic Regression, and AdaBoost also showed highly competitive results with AUCs 0.925, 0.914, and 0.918, respectively, but their performance did not surpass that of XGBoost. ANN demonstrated moderate performance with an AUC of 0.896.

The clear advantage of XGBoost in the test cohort highlights its efficacy in handling the complexities and nuances associated with predicting VAP in TBI patients. This model's ability to maintain high sensitivity and specificity makes it particularly suitable for clinical applications where accurate identification of at-risk patients is crucial. The findings suggest that implementing XGBoost in predictive analytics for healthcare could significantly enhance decision-making processes and patient outcomes.

\subsection{\textit{SHAP analysis}}

\begin{figure*}[!htb]
    \centering
    \includegraphics[width=0.84\linewidth]{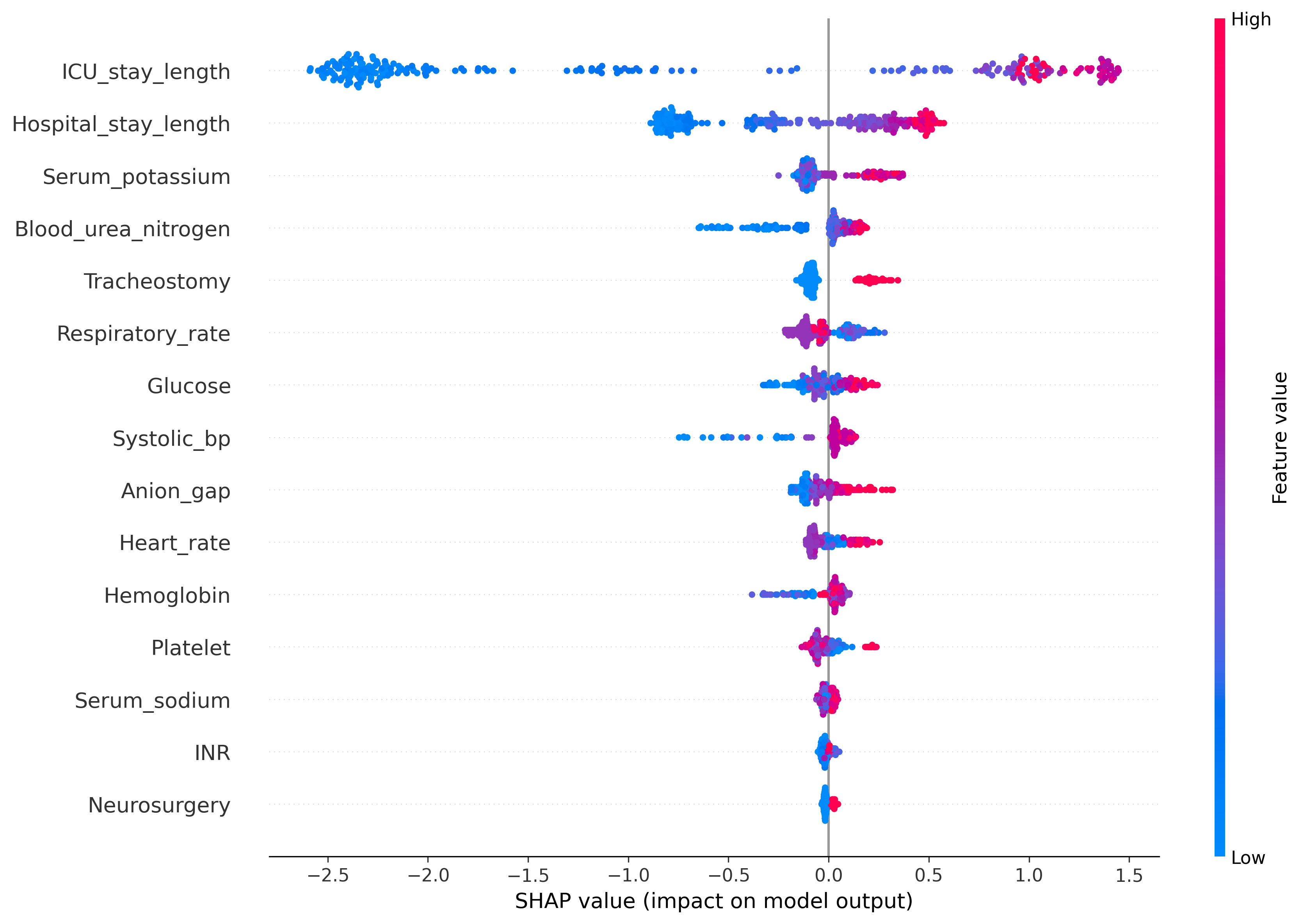}
    \caption{SHAP values for the test set, derived from the XGBoost model.}
    \label{fig:7}
\end{figure*}

The SHAP summary plot is demonstrated in \figref{fig:7}. It provides a detailed understanding of the features impacting the prediction model for VAP in patients suffering from TBI. Notably, ICU length of stay and hospital length of stay are the most influential features as also shown in \cite{ref24}. 

Longer ICU stays significantly decrease the model's predicted output, suggesting a higher risk or poorer outcome, while longer hospital stays also decrease the output but to a lesser extent. High serum potassium levels have a slight positive impact on the model's predictions, whereas high blood urea nitrogen levels show a varied but generally positive effect on the output.

The analysis also reveals that features like respiratory rate and glucose levels, when elevated, positively influence the model's predictions, indicating better outcomes or lower risk. The presence of a tracheostomy shows mixed but generally positive impacts. The non-linear relationships captured by the model highlight the complexity of interactions between features and outcomes. This insight is crucial for improving early detection and interventions for VAP in TBI patients, enhancing model interpretability and clinical applicability. Compared to \cite{ref5}, our SHAP analysis provides a more detailed and nuanced understanding of feature importance. The SHAP summary plot in our study reveals a greater complexity and depth, highlighting the influence of features such as ICU stay length, hospital stay length, serum potassium, and blood urea nitrogen on the model's predictions. This level of detail offers a clearer insight into the factors contributing to VAP development, enhancing the interpretability and reliability of our predictive models.

\section{DISCUSSION}

\subsection{\textit{Summary of Existing Model Compilation}}
In our study, XGBoost demonstrated better performance compared to other models.
This advantage can be explained by the sophisticated mechanisms this model employs. Firstly, XGBoost employs ensemble learning techniques that combine multiple weak learners to form a strong learner, inherently reducing variance and bias for more robust and accurate models. Additionally, these models are particularly effective in handling imbalanced data, a common challenge in medical diagnostics, by optimizing the classification boundary better than simpler models that often struggle with minority class predictions. XGBoost also excels in managing different data types and missing values, and it incorporates built-in regularization to minimize overfitting. XGBoost also incorporates regularization techniques such as L1 (Lasso) and L2 (Ridge) regularization.

However, it is important to note that Logistic Regression showed better performance in terms of specificity and PPV. The specificity of Logistic Regression indicates its higher ability to correctly identify negative cases, reducing the number of false positives, which is crucial in medical diagnostics to avoid unnecessary treatments. Additionally, the higher PPV of Logistic Regression reflects its greater precision in predicting positive cases, which ensures that the identified cases are more likely to be true positives. Nonetheless, since XGBoost performed better in other metrics, we selected it for our proposed model.
  
\subsection{\textit{Comparison with Literature Results}}

A related study by Wang et al. employed the MIMIC-III dataset, utilizing 52 features, significantly more than the 15 features in our analysis \cite{ref5}. This larger number of features likely contributed to severe overfitting in their models on the training set. Their approach included seven different models: SVM, Logistic Regression, Random Forest, Light Gradient Boosting Machine (Light GBM), AdaBoost, Multilayer Perceptron (MLP), and XGBoost. In contrast, we included ANN in our study, which the literature did not utilize. We decided against using Light GBM and MLP after observing suboptimal performance with these models in the original research. The inclusion of ANN was to assess the efficacy of deep learning in our context, although it did not yield particularly satisfactory results.

The previous study provided a table of metrics for each model rather than presenting a single best model. Here, we confirm that our proposed model surpasses every model from the previous study in each metric by comparing our results with the best metrics reported for those models.

For the AUC (Area Under the Curve), our proposed XGBoost model achieved a remarkable 0.940 (95\% CI: 0.935–0.954), significantly surpassing the best AUC of 0.706 (95\% CI: 0.624–0.788) achieved by AdaBoost in the previous study. This substantial improvement of 0.234 in AUC indicates a markedly better discrimination capability of our model. When considering accuracy, our model recorded an impressive 0.875 (95\% CI: 0.861–0.888), far exceeding the 0.640 (95\% CI: 0.616–0.663) accuracy achieved by the previous study's XGBoost model. This demonstrates an improvement of 0.235, reflecting our model's superior ability to correctly predict both positive and negative cases. For sensitivity, our XGBoost model achieved 0.896 (95\% CI: 0.877–0.915), a significant increase over the 0.692 (95\% CI: 0.628–0.756) sensitivity of the previous study's XGBoost model. This improvement of 0.204 highlights our model's enhanced capability to correctly identify true positive cases. In terms of specificity, our model also excelled with a value of 0.857 (95\% CI: 0.828–0.885), slightly improving on the 0.849 (95\% CI: 0.815–0.883) specificity of the previous study's Light GBM model. Although the increase is marginal (0.008), it demonstrates our model's balanced performance in reducing false positives. The Positive Predictive Value (PPV) of our proposed model was 0.792 (95\% CI: 0.761–0.823), which is higher than the 0.708 (95\% CI: 0.615–0.801) PPV achieved by the previous study's Random Forest (RF) model. This improvement of 0.084 indicates better reliability in our model's positive predictions. For Negative Predictive Value (NPV), our model achieved a substantial 0.933 (95\% CI: 0.922–0.944), significantly surpassing the 0.706 (95\% CI: 0.686–0.726) NPV achieved by the previous study's AdaBoost model. This considerable improvement of 0.227 demonstrates our model's strong capability in correctly predicting negative cases. Lastly, for the F1 Score, our proposed XGBoost model achieved 0.842 (95\% CI: 0.826–0.858), significantly higher than the F1 Score of 0.660 (95\% CI: 0.571–0.748) achieved by the previous study's RF model. This improvement of 0.182 underscores our model's balanced precision and recall performance. Additionally, the ROC curves for our models in the test set shown in \figref{fig:6} further substantiate the robustness and generalizability of our approach.

The literature did not employ any method for feature selection and overlooked crucial aspects, such as addressing class imbalance. Our refined feature selection process, which involved eliminating features with high correlations to reduce multicollinearity, and selecting only the top 15 most relevant features based on CatBoost feature importance scores, ensured that our models were trained on the most impactful data. Additionally, we consulted with a clinical expert to verify the relevance of these selected features, further enhancing the robustness and applicability of our models. This approach significantly enhanced the predictive power of our models.

Furthermore, we utilized the SMOTE to address class imbalance, a critical step in medical datasets where the prevalence of conditions can vary significantly. This technique synthesized new examples in the minority class, providing a balanced dataset that improved the generalizability and fairness of our predictive models.

From these comparisons, it is evident that our results are significantly superior to those of the existing literature, particularly in terms of AUC, accuracy, and other related metrics. A higher AUC demonstrates our model's improved accuracy and consistency in distinguishing between patients who will develop VAP and those who will not. This indicates a more reliable predictive capability, crucial in clinical settings to ensure timely and appropriate interventions. Additionally, a higher sensitivity indicates our model's enhanced ability to correctly identify the majority of positive cases, essential in medical diagnostics to avoid missing patients who require urgent care and treatment.

Examining the feature importance derived from our models (shown in \figref{fig:7}), ICU stay length and hospital stay length were the most influential features, followed by serum potassium, blood urea nitrogen, and tracheostomy. In contrast, the literature highlights tracheostomy, RBC transfusion, and percutaneous endoscopic gastrostomy (PEG) as the top features influencing their AdaBoost model. The differences in feature importance underline our model's ability to capture the critical factors impacting VAP development more accurately, enhancing its predictive performance.

The significant improvement in model performance can be primarily attributed to our refined feature selection process and the use of SMOTE. These methodological enhancements are crucial to achieving superior performance metrics compared to the existing literature, highlighting the effectiveness and reliability of our predictive models in clinical applications.

\subsection{\textit{Study Limitations}}

Despite the promising results, there are several limitations in our study that warrant discussion. Firstly, the scope of our data was confined to the MIMIC-III database. Although this dataset is comprehensive, it is over a decade old and may not fully capture the latest clinical practices and patient demographics. To enhance the generalizability of our model, future research should include validation using newer datasets such as MIMIC-IV or other similar repositories. Additionally, expanding the data sources to include text and image data could potentially augment our model’s predictive capabilities, particularly in identifying nuanced clinical indicators of VAP.

Another limitation is the handling of class imbalance using the SMOTE. While this method helped in balancing the training dataset, future studies could investigate more advanced balancing techniques and their impact on model robustness.

Our current model does not account for temporal dependencies in the data, which are crucial for accurately predicting clinical outcomes in an ICU setting. Future work could benefit from incorporating time-series analysis and longitudinal data tracking to better capture patient trajectories and improve predictive accuracy.

Lastly, the practical implementation of our model in clinical settings remains to be tested. Developing user-friendly interfaces and integrating the model into existing electronic health record systems will be essential for real-world application. Collaborating with clinicians during this process can ensure that the tool is both practical and effective in everyday use.

In conclusion, while our study provides a strong foundation for predicting VAP incidence using machine learning, addressing these limitations through future research can significantly enhance the model's accuracy, generalizability, and clinical utility.

\section{CONCLUSION}

From data extraction and cleaning, through feature selection, and resampling, to model choice, hyperparameter selection, and model comparison and tuning, this study has conducted extensive research and effort across many aspects of a complete pipeline to achieve significant improvements in model performance.

These enhancements are notable; we increased the best AUC from 0.706 in the existing literature to 0.940, while also boosting the accuracy from 0.640 to 0.875. This indicates significant advancements in overall prediction, which is crucial for medical data research. Furthermore, most of our baseline models outperformed the previous best study across all metrics, and we successfully reduced the number of features from 52 to 15, streamlining the model without sacrificing performance.

These improvements were primarily achieved through broader feature exploration and selection, the application of data resampling techniques, and meticulous model tuning. 

Given that our research area is currently underexplored and considering that TBI is widespread and imposes a substantial burden on families and society, while VAP frequently occurs in TBI patients, our findings hold substantial medical value. This enhanced performance can lead to better patient management and potentially reduce the mortality associated with complications from VAP, as well as help evaluate the risk of VAP early.

Looking to the future, this work lays a strong foundation for further research. It highlights the potential for using advanced machine learning techniques to tackle similar medical challenges, encourages the exploration of underutilized data within healthcare datasets, and suggests that future studies could focus on refining model accuracy and reliability further, especially in different or larger datasets. This could also inspire the development of real-time predictive tools that could be integrated into clinical workflows, providing timely and accurate support to medical professionals.

\section*{Acknowledgment}

The authors express their appreciation to the team behind MIMIC-III for their invaluable contribution in developing and maintaining such an extensive and detailed public electronic health record (EHR) dataset. This resource has been instrumental in enabling comprehensive research and facilitating significant advancements in the field.

\bibliographystyle{unsrt}
\bibliography{ref}

\end{document}